\documentclass[10pt,twocolumn,letterpaper]{article}

\usepackage{wacv}
\usepackage{times}
\usepackage{epsfig}
\usepackage{graphicx}
\usepackage{amsmath}
\usepackage{amssymb}
\usepackage{booktabs}
\makeatletter
\@namedef{ver@everyshi.sty}{}
\makeatother
\usepackage{tikz}
\usepackage{comment}
\usepackage{amsmath,amssymb} 
\usepackage{color}

\usepackage{graphicx}
\usepackage{booktabs}
\usepackage{microtype}
\usepackage{units}
\usepackage{bbold}
\usepackage{tabularx}
\usepackage{multirow}
\usepackage{pifont}
\newcommand{\cmark}{\ding{51}}%
\newcommand{\xmark}{\ding{55}}%
\usepackage[accsupp]{axessibility}
\def\wacvPaperID{1567} 

\wacvalgorithmstrack   

\wacvfinalcopy 

\ifwacvfinal
\usepackage[breaklinks=true,bookmarks=false]{hyperref}
\else
\usepackage[pagebackref=true,breaklinks=true,colorlinks,bookmarks=false]{hyperref}
\fi

\begin{document}

\title{Is your noise correction noisy? \\ PLS: Robustness to label noise with two stage detection}
\author{Paul Albert, Eric Arazo, Tarun Krishna, Noel E. O'Connor, Kevin McGuinness \\
School of Electronic Engineering, \\
Insight SFI Centre for Data Analytics, Dublin City University (DCU) \\
{\tt\small paul.albert@insight-centre.org}}

\maketitle
\thispagestyle{empty}

\begin{abstract}
   Designing robust algorithms capable of training accurate neural networks on uncurated datasets from the web has been the subject of much research as it reduces the need for time consuming human labor. The focus of many previous research contributions has been on the detection of different types of label noise; however, this paper proposes to improve the correction accuracy of noisy samples once they have been detected. In many state-of-the-art contributions, a two phase approach is adopted where the noisy samples are detected before guessing a corrected pseudo-label in a semi-supervised fashion. The guessed pseudo-labels are then used in the supervised objective without ensuring that the label guess is likely to be correct. This can lead to confirmation bias, which reduces the noise robustness. Here we propose the pseudo-loss, a simple metric that we find to be strongly correlated with pseudo-label correctness on noisy samples. Using the pseudo-loss, we dynamically down weight under-confident pseudo-labels throughout training to avoid confirmation bias and improve the network accuracy. We additionally propose to use a confidence guided contrastive objective that learns robust representation on an interpolated objective between class bound (supervised) for confidently corrected samples and unsupervised representation for under-confident label corrections. Experiments demonstrate the state-of-the-art performance of our Pseudo-Loss Selection (PLS) algorithm on a variety of benchmark datasets including curated data synthetically corrupted with in-distribution and out-of-distribution noise, and two real world web noise datasets. Our experiments are fully reproducible github.com/PaulAlbert31/PLS.
\end{abstract}

\begin{figure}[t]
\centering
\includegraphics[width=\linewidth]{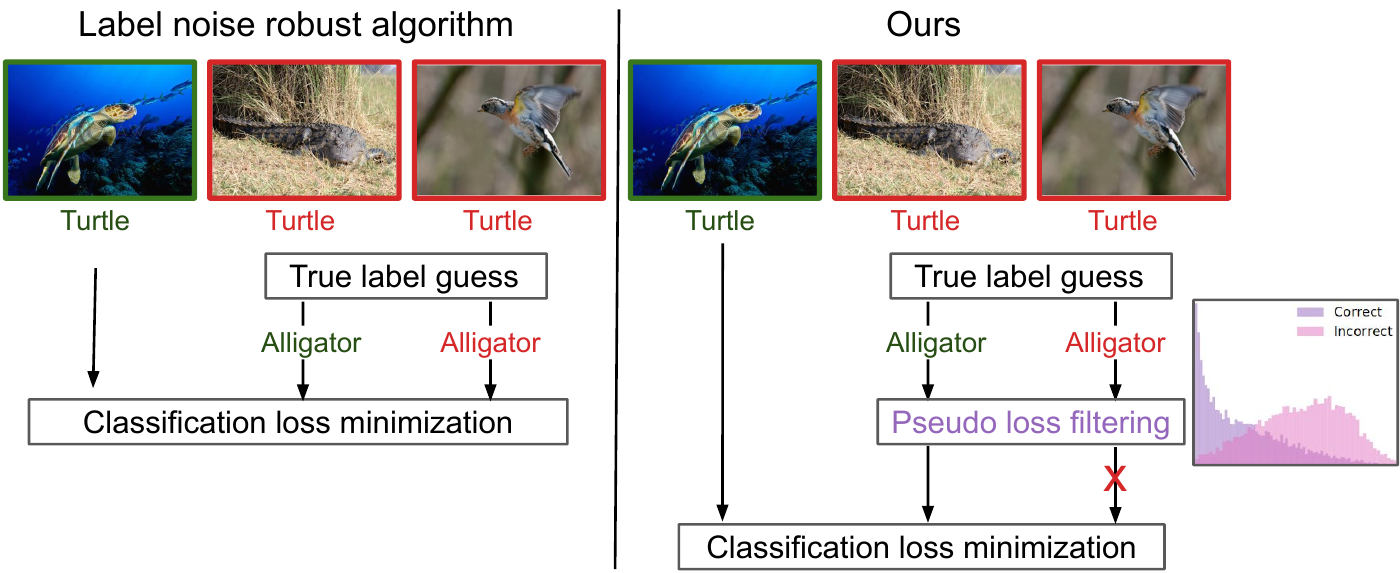}
\par
\caption{Two stage label noise mitigation on detected noisy samples. Contrary to state-of-the-art label noise robust algorithms, we filter out incorrect pseudo-labels using the pseudo-loss to avoid confirmation bias on incorrect corrections.~\label{fig:contrib}}
\end{figure}
\section{Introduction}
Standard supervised datasets for image classification using deep learning~\cite{2012_NeurIPS_ImageNet,2011_PASMCL_pascalVOC,2014_ECCV_COCO,2009_CIFAR} are constituted by large amounts of images gathered from the web which have been heavily curated by multiple human annotators. In this paper, we propose to devise an algorithm which aims to train an accurate classification network on a web crawled dataset~\cite{2017_arXiv_WebVision,2021_ICCV_weblyfinegrained} where the human curation process was skipped. By doing so, the dataset creation time is greatly reduced but label noise becomes an issue~\cite{2022_WACV_DSOS} and can greatly degrade the classification accuracy~\cite{2017_ICLR_Rethinking}. To counter the effect of noisy annotations, previous contributions have focused on detecting the noisy samples using the natural robustness of deep learning architectures to noise in early training stages~\cite{2019_ICML_BynamicBootstrapping,2017_ICML_Memorization}. These algorithms will identify noisy samples because they tend to be learned slower than their clean counterpart~\cite{2020_ICLR_DivideMix}, because of incoherences with the labels of close neighbors in the feature space~\cite{2021_CVPR_MOIT,2021_ICCV_RRL}, a confident prediction from the neural net in a different class than the target class~\cite{2021_CVPR_JoSRC,2020_NeurIPS_EarlyReg}, inconsistent predictions across iterations~\cite{2020_ICPR_Robust,2019_ICLR_Forgetting}, and more. Once the noisy samples are identified, a corrected label is produced, yet ensuring that labels are correctly guessed is less studied in the label noise literature. Some propositions inspired by semi-supervised learning~\cite{2020_arXiv_FixMatch,2021_NeurIPS_flexmatch} have been made recently by Li~\etal~\cite{2021_ICCV_RRL} where only pseudo-labels whose value in the max softmax bin (confidence) is superior to a hyper-parameter threshold value are kept or by Song~\etal~\cite{2019_ICML_SELFIE} where low entropy predictions indicate a confident pseudo-label. This paper proposes to focus on the correction of noisy samples once they have been detected. We specifically propose a novel metric, the pseudo-loss, which is able to retrieve correctly guessed pseudo-labels and that we show to be superior to the pseudo-label confidence previously used in the semi-supervised literature. We find that incorrectly guessed pseudo-labels are especially damaging to the supervised contrastive objectives that have been used in recent contributions~\cite{2021_CVPR_MOIT,2022_ECCV_SNCF,2021_ICCV_RRL}. We propose an interpolated contrastive objective between class-conditional (supervised) for the clean or correctly corrected samples, where we encourage the network to learn similar representation for images belonging to the same class; and an unsupervised objective for the incorrectly corrected noise. This results in {\b P}seudo-{\b L}oss {\b S}election (PLS) a two-stage noise detection algorithm where the first stage detects all noisy samples in the dataset while the second stage removes incorrect corrections. We then train a neural network to jointly minimize a classification and a supervised contrastive objective. We design PLS on synthetically corrupted datasets and validate our findings on two real world noisy web crawled datasets. Figure~\ref{fig:contrib} illustrates our proposed improvement to label noise robust algorithms. Our contributions are:

\begin{itemize}
    \item A two-stage noise detection using a novel metric where we ensure that the corrected targets for noisy samples are accurate;
    \item A novel softly interpolated confidence guided contrastive loss term between supervised and unsupervised objective to learn robust features from all images;
    \item Extensive experiments of synthetically corrupted and web-crawled noisy datasets to demonstrate the performance of our algorithm.
\end{itemize}

\section{Related work}
\subsection*{Label noise robust algorithms}
Label noise in web crawled datasets has been evidenced to be a mixture between in-distribution (ID) noise and out-of-distribution (OOD) noise~\cite{2022_WACV_DSOS}. In-distribution noise denotes an image that was assigned an incorrect label but can be corrected to another label in the label distribution. Out-of-distribution noise are images whose true label lie outside of the label distribution and cannot be directly corrected. While some algorithms have been designed to detect ID and OOD separately, others reach good results by assuming all noise is ID. The rest of this section will introduce state-of-the-art approaches to detect and correct noisy samples. 

\subsection{Label noise detection} 
Label noise in datasets can be detected by exacerbating the natural resistance of neural networks to noise. Small loss algorithms~\cite{2019_ICML_BynamicBootstrapping,2020_ICLR_DivideMix,2020_ICPR_Robust} observe that noisy samples tend to be learned slower than their clean counterparts and that a bi-modal distribution can be observed in the training loss where noisy samples belong to the high loss mode. A mixture model is then fit to the loss distribution to retrieve the two modes in an unsupervised manner. Other approaches evaluate the neighbor coherence in the network feature space where images are expected to have many neighbors from the same class~\cite{2021_CVPR_MOIT,2021_ICCV_RRL,2021_arXiv_Scanmix} and a hyper-parameter threshold is used on the number of neighbors from the same class to allow to identify the noisy samples. In some cases, a separate OOD detection can be performed to differentiate between correctable ID noise and uncorrectable OOD samples. OOD samples are detected by evaluating the uncertainty of the current neural network prediction. EvidentialMix~\cite{2020_WACV_EDM} uses the evidential loss~\cite{2018_NeurIPS_evidentialloss}, JoSRC evaluates the Jensen-Shannon divergence between predictions~\cite{2021_CVPR_JoSRC}, and DSOS~\cite{2022_WACV_DSOS} computes the collision entropy.
An alternative approach is to use a clean subset to learn to detect label noise in a meta-learning fashion~\cite{2020_arXiv_MetaSoftApple,2018_NeurIPS_usingtrusted,2017_NeurIPS_Towards,2021_CVPR_FaMUS} but we will assume in this paper that a trusted set is unavailable.

\subsection{Noise correction}
Once the noisy samples have been detected, state-of-the-art approaches guess true labels using current knowledge learned by the network. Options include guessing using the prediction of the network on unaugmented samples~\cite{2019_ICML_BynamicBootstrapping,2020_NeurIPS_EarlyReg}, semi-supervised learning~\cite{2020_ICLR_DivideMix,2021_CVPR_MOIT}, or neighboring samples in the feature space~\cite{2021_ICCV_RRL}. Some approaches also simply discard the detected noisy examples to train on the clean data alone~\cite{2020_ICML_MentorMix,2018_ICML_MentorNet,2019_ICML_Iterativetrimmed,2019_ICML_disagreementweights}. In the case where a separate out-of-distribution detection is performed, the samples can either be removed from the dataset~\cite{2020_WACV_EDM}, assigned a uniform label distribution over the classes to promote rejection by the network~\cite{2021_CVPR_JoSRC,2022_WACV_DSOS}, or used in an unsupervised objective~\cite{2022_ECCV_SNCF}.

\subsection{Noise regularization}
Another strategy when training on label noise datasets is to use strong regularization either in the form of data augmentation such as mixup~\cite{2018_ICLR_Mixup} or using a dedicated loss term~\cite{2020_NeurIPS_EarlyReg}. Unsupervised regularization has also shown to help improve the classification accuracy of neural networks trained on label noise datasets~\cite{2021_ICCV_RRL,2021_TM_COLDL}.

\section{PLS}
We consider an image dataset dataset $\mathcal{X} = \{x_i\}_{i=1}^N$ associated with one-hot encoded classification labels $\mathcal{Y}$ over $C$ classes. An unknown percentage of labels in $\mathcal{Y} = \{y_i\}_{i=1}^N$ are noisy,~\ie $y_i$ is different from the true label of $x_{i}$. We aim to train a neural network $\phi$ on the imperfect label noise dataset to perform accurate classification on a held out test set.

\subsection{Detecting the noisy samples~\label{sec:noisedetect}}
Our contributions do not include detecting the noisy labels but we propose to focus here on improving the correction of the noisy samples once they have been detected. We use a known phenomenon in previous research for label noise classification~\cite{2019_ICML_BynamicBootstrapping,2020_ICLR_DivideMix,2020_ICPR_Robust} where in early stages of training, the cross-entropy loss between $\phi$'s prediction on an unaugmented view on an image $\phi(x_i)$ and the associated (possibly noisy) ground-truth label $y_i$ is observed to separate into a low loss clean mode and high loss noisy mode. We therefore propose to fit a Gaussian Mixture Model (GMM) to the training loss to retrieve each mode in an unsupervised fashion. Clean samples are finally identified as belonging to the low loss mode with a probability superior to a threshold $t=0.95$. Alternative metrics have been proposed to retrieve noisy labels but we find that while approaches retrieve noisy samples very similarly for synthetic noise, the training loss is more accurate in the case of real world noise. We justify this statement in Section~\ref{sec:expret}.

\subsection{Confident correction of noisy labels}
\subsubsection{Guessing labels for detected noisy samples~\label{sec:labelguessing}}
To guess the true label of detected noisy samples, we propose to use a consistency regularization approach. Given an image $x_i$ associated to a noisy label, we produce two weakly augmented views $x_{i1}$ and $x_{i2}$. Weak augmentations are random cropping after zero-padding and random horizontal flipping. Using the current state of $\phi$, we guess the pseudo-label $\hat{y}_i$ as
\begin{equation}
    \hat{y}_i = \left(\frac{\phi(x_{i1})+\phi(x_{i2})}{2}\right)^\gamma,
\end{equation}
with $\gamma = 2$ being a temperature hyper-parameter. We then apply a max normalization over $\hat{y}_i$ to ensure that the values of the pseudo-label are between 0 and 1.

\subsubsection{Correcting only confident pseudo-labels}
We propose to only correct those pseudo-labels that are likely to be correctly guessed by $\phi$. This solution has already been explored in the semi-supervised literature~\cite{2020_arXiv_FixMatch,2021_NeurIPS_flexmatch} where pseudo-labels are only kept if the value of the maximum probability is superior to an hyperparameter threshold. Both prediction confidence measured by highest probability bin~\cite{2021_ICCV_RRL} or prediction entropy~\cite{2019_ICML_SELFIE} has also been successfully applied in the label noise literature. We propose to identify correct pseudo-labels by evaluating a different metric, which we name the pseudo-loss. The pseudo-loss evaluates the cross-entropy loss between the pseudo-label $\hat{y}_i$  and the prediction of the model on an unaugmented view $\phi(x_i)$:
\begin{equation}
    l_\text{pseudo} = - \hat{y}_i  \log \phi(x_i).
\end{equation}
We observe that much like the noise detection loss in Section~\ref{sec:noisedetect}, the pseudo-loss is bi-modal (see Figure~\ref{fig:contrib} and Section~\ref{sec:exppseudo}). We propose to fit a second GMM to the pseudo-loss and to use the posterior probability of a sample to belong to the low $l_{pseudo}$ mode (correct pseudo-label, left-most gaussian) as $w$, a weight in the classification loss $l_{classif}$ that reduces the impact of incorrect pseudo-labels. Underconfident, high pseudo-loss samples are weighed with values close to 0 (low probability of belonging to the low pseudo-loss mode) while confident pseudo-labels are weighed with values close to 1 (high probability of belonging to the low pseudo-loss mode).
The classification loss we use is a weighed cross-entropy with mixup:
\begin{equation}
    l_\text{classif} = \frac{1}{\sum_{i=1}^N w_{mix,i}} \sum_{i=1}^N -w_{mix, i}\hat{y}_{mix,i} \log \phi(x_{mix,i}),
    \label{eq:pseudolossce}
\end{equation}
where $w_{mix}$, $x_{mix}$ and $y_{mix}$ are linearly interpolated with another random sample in the mini-batch using parameter $\lambda \sim \mathcal{U}(0,1)$, sampled for every mini-batch (mixup~\cite{2018_ICLR_Mixup}). We evaluate how the pseudo-loss compares to pseudo-label confidence in Section~\ref{sec:exppseudo}.

\subsubsection{Supervised contrastive learning}
To improve the quality of representations learned by $\phi$, we propose to train a supervised contrastive objective jointly with the classification loss. We compute the contrastive features as a linear projection $g$ from the classification features to the $L_2$ normalized contrastive space. A contrastive objective aims to learn similar contrastive features for images belonging to the same class. Given a training mini-batch of images $X_b$ with associated classification labels $Y_b$, we produce a weakly augmented view $X_{b1}$ and a strongly augmented view $X_b'$. The strong augmentations are the SimCLR augs~\cite{2020_arXiv_SimCLRv2}: random resized crop, color jitter, random grayscale, and random horizontal flipping. We compute the label similarity matrix $L = Y_bY_b^t$ and the feature similarity matrix:
\begin{equation}
P = \frac{g(\phi(X_{i1})) g(\phi(X_i'))^T}{\mu},
\end{equation} with $\mu=0.2$ being a temperature scaling parameter. Both $P$ and $L$ are $B\times B$ matrices with $B$ the mini-batch size. The contrastive loss is the row-wise cross-entropy loss: 
\begin{equation}
    l_\text{naivecont} = \frac{1}{B} \sum_{i=1}^B - \frac{L_{i} \log P_i}{\sum_{c=1}^C L_{i,c}},
\end{equation}
where $L_i$ and $P_i$ denote the row $i$ of the corresponding matrix.
Because label noise is present in the datasets we train on, minimizing $l_\text{naivecont}$ directly is detrimental since similarities will be enforced between samples whose pseudo-label cannot be trusted. We propose instead to account for pseudo-label incorrectness and train a confidence guided contrastive objective.

\begin{figure*}[t]
\centering
\includegraphics[width=.7\linewidth]{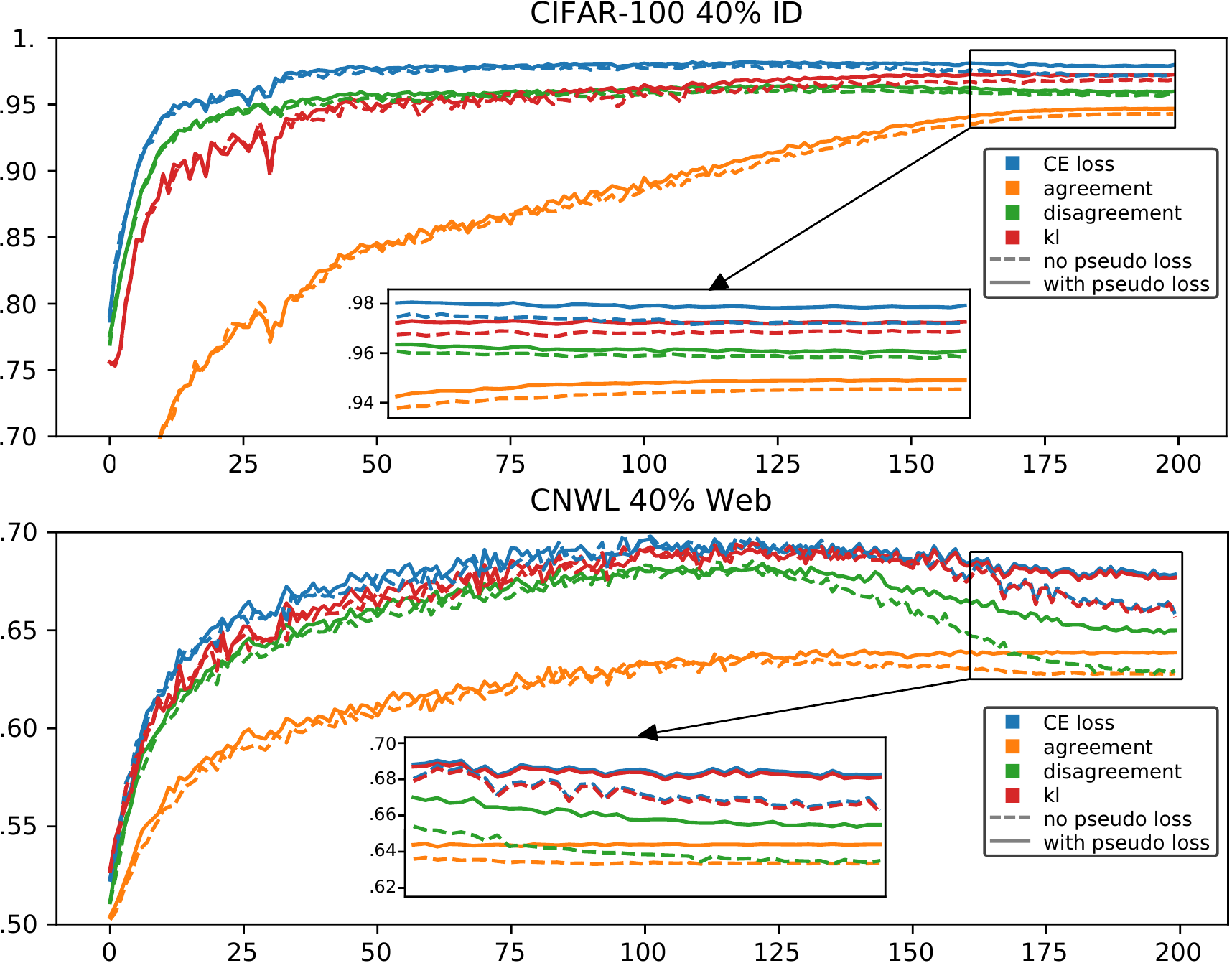}
\par
\caption{AUC for commonly used metrics in the literature when retrieving noisy samples. Full lines indicate that we remove incorrect pseudo-labels with the pseudo-loss. Dashed lines indicate that all pseudo-labels are used. Accounting for incorrect pseudo-labels improves the detection of noisy samples. ~\label{fig:retnoise}}
\end{figure*}

\begin{figure}[t]
\centering
\includegraphics[width=\linewidth]{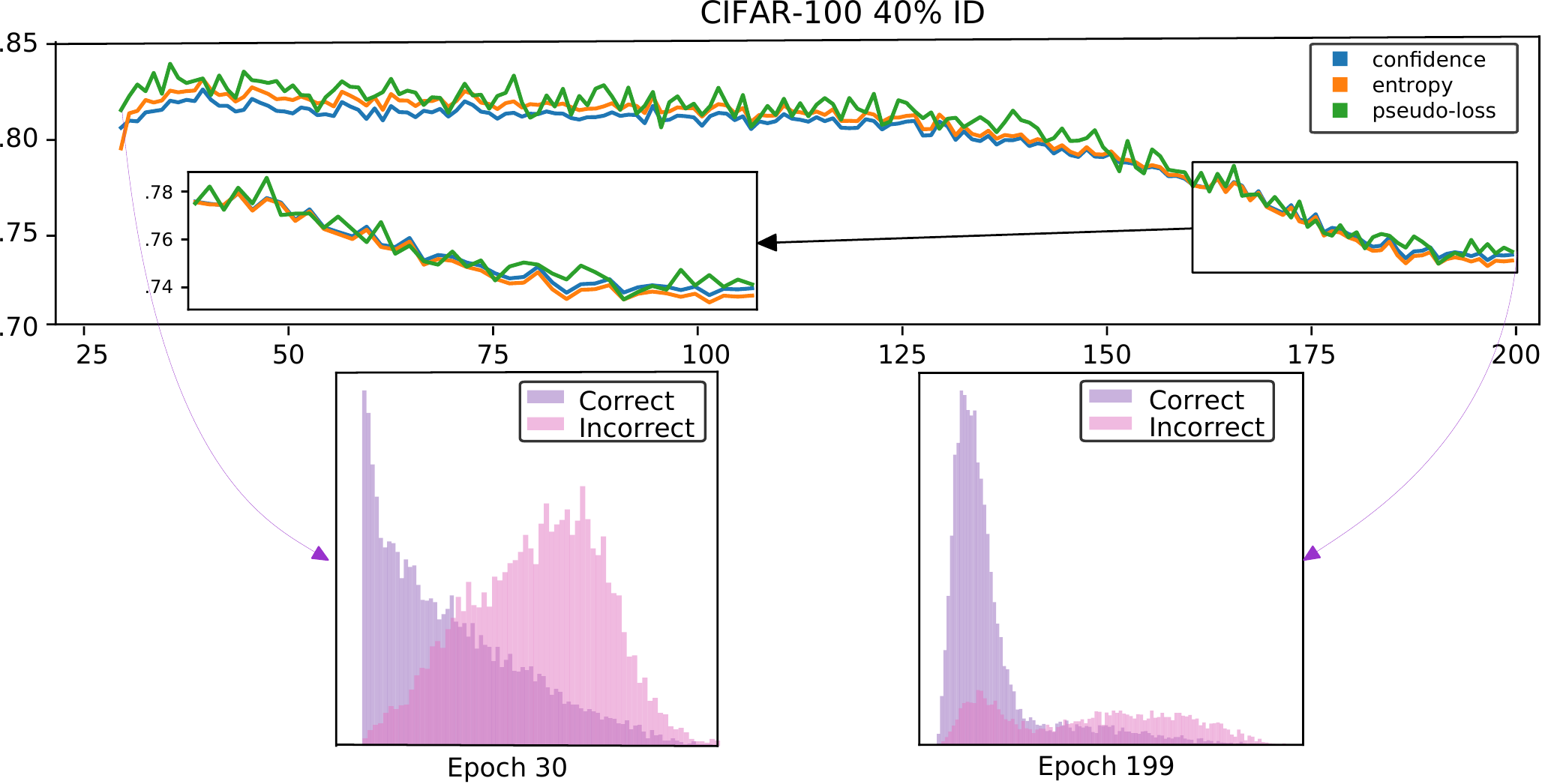}
\par
\caption{Pseudo-loss bi-modality and capacity to retrieve correctly guessed pseudo-labels. The top graph represents the AUC when retreiving correctly guessed pseudo-labels using the prediction confidence, entropy or the pseudo-loss. The bottom part of the figure shows the bi-modality of the pseudo-loss at two points during training and the confirmation bias at the end of the training.~\label{fig:retpseudo}}
\end{figure}

\begin{figure}[t]
\centering
\includegraphics[width=\linewidth]{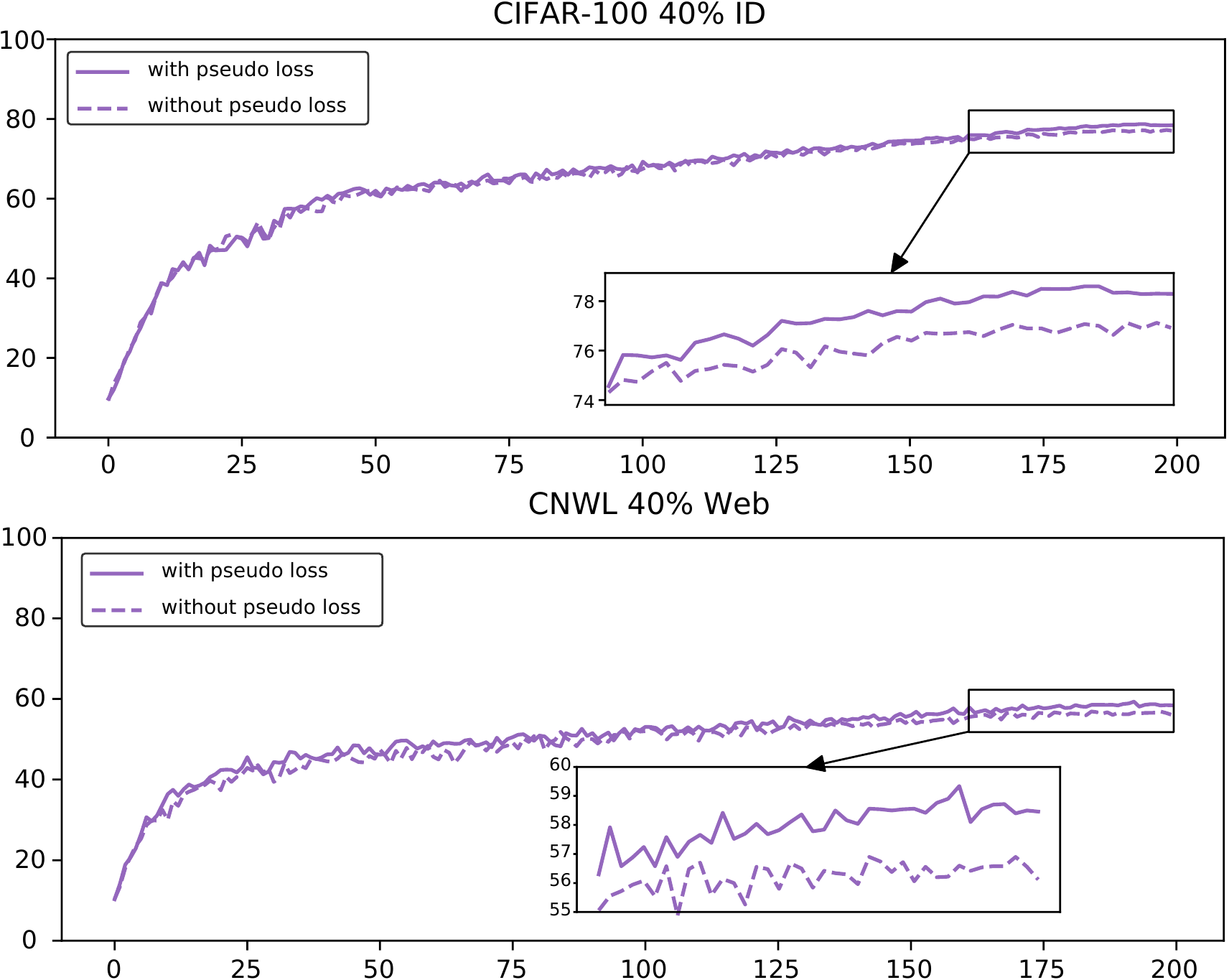}
\par
\caption{Validation accuracy per epoch, with or without removing incorrect pseudo-labels using the pseudo-loss~\label{fig:noiseacc}}
\end{figure}

\subsubsection{Confidence guided contrastive learning in the presence of label noise}
So far the proposed confidence guided contrastive objective does not account for label noise in the dataset which conflicts with the classification objective and will be harmful to the learned representation. The first step to account for label noise is to replace the correctly guessed labels in Section~\ref{sec:labelguessing} for the detected noisy samples to produce $\hat{Y}_b$. As we do for the classification loss, we also want to prevent incorrect pseudo-labels from interfering with the contrastive algorithm. Rather than simply weighting the contrastive loss for a low confidence pseudo-labeled noisy sample, we propose to use the unsupervised capabilities of contrastive losses. Depending on the confidence in a pseudo-label for a noisy sample, the pseudo-label will be used to enforce similar features with other samples from the same guessed class (high pseudo-label confidence) or will only be encouraged to learn similar features between augmented views of the same image (low confidence pseudo-label). To do so in a continuous manner, without the need for a threshold on $w$, we modify the initial classification labels $\hat{Y}_b$ by concatenating a weighted one-hot positional encoding of samples in the mini-batch with $\hat{Y}_b$ using $w$. The label for sample $i$ in the mini-batch becomes: 
\begin{equation}
    y_{\text{cont},i} = \text{concat}(w_{i}\times \hat{y}_i, (1-w_{i})\times\mathcal{O}(i,B)),
    \label{eq:pseudolosscont}
\end{equation}
where $\mathcal{O}(i,B)$, the one-hot positional encoding of the sample $i$ in the mini-batch is a zero-vector of size $B$ with value $1$ at position $i$ with $B$ the mini-batch size. An illustration for computing $y_{cont}$ is available in the supplementary material. Repeating the process, we create $Y_{cont,b}$ of size $B\times (C+B)$. Finally, to benefit from the noise robustness of mixup in the contrastive objective, we adopt a similar setting as in iMix~\cite{2021_ICLR_iMix} and linearly interpolate $X_{i1}$ among samples in the mini-batch to create $X_{mix,i}$ (InputMix) before extracting the features as well as the corresponding label $Y_{cont_b}$ to create $Y_{mix_b}$ using $\beta \sim \mathcal{U}(0,1)$.
To compute the confidence guided contrastive objective, we use $L' = Y_{mix_b}Y_{mix_b}^t$ and $P' =  g(\phi(X_{mix,i}))g(\phi(X_i'))^T/\mu$. The confidence guided, noise robust contrastive learning loss we minimize is
\begin{equation}
    l_\text{cont} = \frac{1}{B} \sum_{i=1}^B - \frac{L'_{i} \log P'_i}{\sum_{c=1}^C L'_{i,c}},
\end{equation}
The final training objective we optimize is: 
\begin{equation}
    l = l_\text{classif} + l_\text{cont}.
\end{equation}

\section{Experiments}
\subsection{Setup~\label{sec:setup}}
We conduct noise robustness experiments on four image datasets. For synthetically corrupted datasets, we train on CIFAR-100 and miniImageNet. CIFAR-100 is corrupted with symmetric or asymmetric in-distribution noise where we randomly flip a the labels of a fixed percentage of the dataset to another from the same distribution. For out-of-distribution noise, we replace a fixed percentage of samples with images from ImageNet32 or Places365 as in Albert~\etal~\cite{2022_ECCV_SNCF} where $r_{in}$ and $r_{out}$ respectively denotes the in-distribution and out-of-distribution noise ratios. For miniImageNet, we use the web noise corruption from Jiang~\etal~\cite{2020_ICML_MentorMix}. We train on both of these datasets at a resolution of $32\times 32$ with a pre-activation ResNet18~\cite{2016_CVPR_ResNet}. We train for $200$ epochs, starting with a learning rate of $0.1$. We use a batch size of $256$, stochastic gradient descent with a weight decay of $5\times 10^{-5}$. We perform the warmup phase with the supervised objective only for $30$ epochs on CIFAR-100 and for $1$ epoch on MiniImageNet. We evaluate our approach on real world data by conducting experiments on the webly fined grained datasets~\cite{2021_ICCV_weblyfinegrained}. We follow the setup of Zeren~\etal~\cite{2021_TM_COLDL} and use a ResNet50~\cite{2016_CVPR_ResNet} pretrained on ImageNet~\cite{2012_NeurIPS_ImageNet} at a resolution of $448 \times 448$. We train with a learning rate of $0.003$ and use a batch size of  $16$, stochastic gradient descent with a weight decay of $10^{-3}$ and warmup for $10$ epochs. We find that the class-balanced regularization (class reg) commonly used in the label noise litterature~\cite{2020_ICLR_DivideMix,2021_CVPR_MOIT} helps improve the validation accuracy so we minimize it jointly to the classification objective. For all experiments, we employ a cosine learning rate decay after the end of the warmup phase.

\subsection{Detecting and correcting label noise~\label{sec:expret}}
We propose to evaluate how commonly used metrics fair at retrieving noisy samples in both synthetic and controlled web noisy data. Figure~\ref{fig:retnoise} plots AUC retrieval scores for noisy samples for different metrics proposed in the literature. We study neighbor agreement and disagreement in the contrastive feature space as in~\cite{2021_CVPR_MOIT} which is also used in~\cite{2021_ICCV_RRL}; training (small) loss based methods as in~\cite{2019_ICML_BynamicBootstrapping,2020_ICLR_DivideMix}; Kullback–Leibler (kl) divergence as in~\cite{2021_CVPR_JoSRC,2022_CVPR_PNP,2021_TM_COLDL}. Area under the curve (AUC) retreival score for the noisy samples are reported at every epoch in Figure~\ref{fig:retnoise}. We observe that metrics behave similarly in the presence of ID noise but greater differences are observed when retrieving controlled web noise from the CNWL dataset. We observe in that case that the cross-entropy loss (small loss) is the most accurate at retrieving noisy web samples. Note also that the retrieval accuracy of the different metrics greatly drops when compared to the synthetic in-distribution noise which prompts further research to improve the detection of web noise when no held-out labeled set is present. 

\begin{table}[t]
    \caption{Ablation study and importance of the pseudo-loss selection $w$ for the contrastive loss $l_{cont}$. Experiments conducted on CIFAR-100 corrupted with $r_{in} = 0.4$, $r_{in} = r_{out} = 0.2$ and on the CNWL corrupted with 40\% web noise. Top-1 accuracy.
    \label{tab:ablacont}}
    \global\long\def\arraystretch{0.9}%
    \centering
    \resizebox{.95\linewidth}{!}{%
    \centering
    \begin{tabular}{c>{\centering}c>{\centering}c>{\centering}c>{\centering}c>{\centering}c>{\centering}c>{\centering}c>{\centering}c>{\centering}c>{\centering}c>{\centering}c>{\centering}c}
    \toprule
    & & & \multicolumn{3}{c}{Noise level}\tabularnewline
    \cmidrule{4-6}
    & & & & $r_{in} = 0.2$ & \tabularnewline
    correct & cont & $w$ &  $r_{in} = 0.4$ & $r_{out} = 0.4$ & CNWL $40\%$ \tabularnewline
    \midrule
    \xmark & \xmark & \xmark & $66.31$ & $59.54$ & $46.40$ \tabularnewline
    \cmark & \xmark & \xmark & $74.53$ & $69.17$ & $55.08$ \tabularnewline
    \cmark & \xmark & \cmark & $75.57$ & $69.53$ & $55.72$ \tabularnewline
    \cmark & \cmark & \xmark & $76.21$ & $70.10$ & $58.62$ \tabularnewline
    \cmark & \cmark & \cmark & $77.43$ & $72.21$ & $59.90$ \tabularnewline
    \midrule
    \cmark & \cmark & $l_{classif}$ & $77.84$ & $69.09$ & $57.76$ \tabularnewline
    \bottomrule
    \end{tabular}}
\end{table}

\begin{table*}[t]
    \caption{Mitigating ID noise on CIFAR-100. Accuracy numbers from respective papers or run using publicly available code. We bold the highest best accuracy and report standard deviation over 3 random noisy datasets and network initializations.
    \label{tab:sotac100in}}
    \global\long\def\arraystretch{0.8}%
    \centering
    \resizebox{.8\textwidth}{!}{%
    \centering
    \begin{tabular}{c>{\centering}c>{\centering}c>{\centering}c>{\centering}c>{\centering}c>{\centering}c>{\centering}c>{\centering}c>{\centering}c>{\centering}c>{\centering}c>{\centering}c}
    \toprule
    Noise type & $r_{in}$ & CE & M & DB & DM & ELR+ & MOIT+ & Sel-CL+ & RRL  & PLS \tabularnewline
    \midrule
    \multirow{4}{*}{Symetric} & 0.0 & 76.99 & 79.29 & 64.79 & 72.75 & 83.14 & 77.07 & 79.90 & $\textbf{80.70}$ & $78.85 \pm 0.21$ \tabularnewline
    & 0.2 & 62.60 & 71.55 & 73.9 & 77.3 & 77.6 & 75.89 & 76.5 & $79.4 \pm 0.1$ &  $\textbf{80.03} \pm 0.15$ \tabularnewline
    & 0.5 & 46.59 & 61.12 & 66.1 & 74.6 & 73.6 & 67.54 & 72.4 & $75.0 \pm 0.4$ & $\textbf{76.48} \pm 0.25$ \tabularnewline
    & 0.8 & 23.46 & 37.66 & 45.67 & 60.2 & 60.08 & 51.36 & 59.6 & 32.21 & $\textbf{63.33} \pm 0.38$\tabularnewline
    \bottomrule
    \end{tabular}}
    
\end{table*}

\subsection{Identifying incorrect pseudo-labels~\label{sec:exppseudo}}
We aim to demonstrate that not accounting for pseudo-label correctness in the detected noise is detrimental to both noise detection in the subsequent epochs as well as the overall true label recovery and validation accuracy throughout the training. We compare the pseudo-loss against the prediction confidence or entropy of the pseudo-label which commonly in the semi-supervised literature in Figure~\ref{fig:retpseudo} where we train on CIFAR-100 corrupted with 40\% of ID noise with no pseudo-label filtering. We observe that the pseudo-loss is on par with other metrics when retrieving correctly guessed pseudo-labels and that the detection on incorrect pseudo-labels becomes more challenging throughout the training as the learning rate is reduced and confirmation bias increases. More importantly, we find the pseudo-loss distribution over detected noisy samples to be bi-modal much like the small loss for noise detection. Consequently, we apply the same methodology as for the first stage detection where we fit a two mode gaussian mixture to the pseudo-loss and use the probability of a sample to belong to the low loss (correct pseudo-label) mode as $w$. This allows us to remove the need for a hyper-parameter threshold on the pseudo-label confidence/entropy as is always the case in the semi-supervised literature. PLS dynamically adapts as the training progresses and the network becomes naturally very confident in its predictions.

\subsection{Pseudo-loss based selection of correct pseudo-labels}
After assigning a probability of being correct for all guessed pseudo-labels on detected noisy samples, we evaluate how correct pseudo-label selection using the pseudo-loss influences label noise detection when the correction starts in Figure~\ref{fig:retnoise}. We train a neural network with or without our proposed pseudo-label selection using the pseudo-loss weights from equations~\ref{eq:pseudolossce} and~\ref{eq:pseudolosscont} (full line). We observe that both the noise retrieval and validation accuracy (Figure~\ref{fig:noiseacc}) are improved when incorrect pseudo-labels are removed. By avoiding to compute weight updates over incorrect pseudo-labels, our pseudo-loss selection reduces confirmation bias and improves the retrieval accuracy of noisy samples. Treating un-guessable samples as unlabeled in equation~\ref{eq:pseudolosscont} helps us to further improve the classification accuracy (see ablation study in Table~\ref{tab:ablacont}).

\setlength{\tabcolsep}{3pt}
\begin{table*}[t]
    \caption{Mitigating ID noise and OOD noise on CIFAR-100 corrupted with ImageNet32 or Places365 images. Accuracy numbers from~\cite{2022_ECCV_SNCF}. We bold the highest best accuracy and report standard deviation over 3 random noisy corruptions and network initialization.
    \label{tab:sotac100inout}}
    \global\long\def\arraystretch{0.8}%
    \centering
    \resizebox{.8\textwidth}{!}{%
    \centering
    \begin{tabular}{l>{\centering}c>{\centering}c>{\centering}c>{\centering}c>{\centering}c>{\centering}c>{\centering}c>{\centering}c>{\centering}c>{\centering}c>{\centering}c>{\centering}c}
    \toprule
    Corruption & $r_{out}$ & $r_{in}$ & CE & M & DB & JoSRC & ELR & EDM & {DSOS} & RRL & SNCF & PLS \tabularnewline
    \midrule
    \multirow{4}{*}{INet32}& $0.2$ & $0.2$ & $63.68$ & $66.71$ & $65.61$ & $67.37$ & $68.71$ & $71.03$ & $70.54$ & $72.64$ & $72.95$ & $\textbf{76.29} \pm 0.28$ \tabularnewline
    & $0.4$ & $0.2$ & $58.94$ & $59.54$ & $54.79$ & $61.70$ & $63.21$ & $61.89$ & $62.49$ & $66.04$ & $67.62$ & $\textbf{72.06} \pm 0.19$ \tabularnewline
    & $0.6$ & $0.2$ & $46.02$ & $42.87$ & $42.50$ & $37.95$ & $44.79$ & $21.88$ & $49.98$ & $26.76$ & $53.26$ & $\textbf{57.78} \pm 0.26$  \tabularnewline
    & $0.4$ & $0.4$ & $41.39$ & $38.37$ & $35.90$ & $41.53$ & $34.82$ & $24.15$ & $43.69$ & $31.29$ & $54.04$ & $\textbf{56.92} \pm 0.49$ \tabularnewline
    \midrule
    \multirow{4}{*}{Places365}& $0.2$ & $0.2$ & $59.88$ & $66.31$ & $65.85$ & $67.06$ & $68.58$ & $70.46$ & $69.72$ & $72.62$ & $71.25$ & $\textbf{76.35} \pm 0.05$ \tabularnewline
    & $0.4$ & $0.2$ & $53.46$ & $59.75$ & $55.81$ & $60.83$ & $62.66$ & $61.80$ & $59.47$ & $65.82$ & $64.03$ & $\textbf{71.65} \pm 0.61$ \tabularnewline
    & $0.6$ & $0.2$ & $39.55$ & $ 39.17$ & $40.75$ & $39.83$ & $37.10$ & $23.67$ & $35.48$ & $49.27$ & $49.83$ & $\textbf{57.31} \pm 0.31$ \tabularnewline
    & $0.4$ & $0.4$ & $32.06$ & $34.36$ & $35.05$ & $33.23$ & $34.71$ & $20.33$ & $29.54$ & $26.67$ & $50.95$ & $\textbf{55.61} \pm 0.55$ \tabularnewline
       \bottomrule
    \end{tabular}}
    
\end{table*}

\begin{table}[t]
    \centering
    \caption{Comparison against state-of-the-art algorithms on the fine grained web datasets. We bold the best results. Top-1 best accuracy.\label{tab:sotawebfg}}
    \global\long\def\arraystretch{0.8}%
    \resizebox{.75\linewidth}{!}{{{}}%
    \begin{tabular}{l>{\centering}c>{\centering}c>{\centering}c>{\centering}c>{\centering}c}
    \toprule
       Algorithm & Web-Aircraft & Web-bird & Web-car \tabularnewline
       \midrule
       CE & 60.80 & 64.40 & 60.60 \tabularnewline
       Co-teaching & 79.54 & 76.68 & 84.95 \tabularnewline
       PENCIL & 78.82 & 75.09 & 81.68 \tabularnewline
       SELFIE & 79.27 & 77.20 & 82.90 \tabularnewline 
       DivideMix & 82.48 & 74.40 & 84.27 \tabularnewline
       Peer-learning & 78.64 & 75.37 & 82.48 \tabularnewline
       PLC & 79.24 & 76.22 & 81.87 \tabularnewline
       \midrule
       PLS & \textbf{87.58} & \textbf{79.00} & \textbf{86.27} \tabularnewline
    \bottomrule
    \end{tabular}}
\end{table}

\begin{table*}[t]
    \caption{Web-corrupted miniImageNet from the CNWL~\cite{2020_ICML_MentorMix} ($32\times 32$). We run our algorithm; other results are from~\cite{2021_arXiv_propmix}. We report top-1 best accuracy and bold the best results.
    \label{tab:sotamini32}}
    \global\long\def\arraystretch{0.9}%
    \centering
    \resizebox{.65\textwidth}{!}{%
    \centering
    \begin{tabular}{c>{\centering}c>{\centering}c>{\centering}c>{\centering}c>{\centering}c>{\centering}c>{\centering}c>{\centering}c>{\centering}c>{\centering}c}
    \toprule
    Noise level & CE & M & DM & MM & FaMUS & SM & PM & SNCF & PLS \tabularnewline
    \midrule
    20 & $47.36$ & $49.10$ & $50.96$ & $51.02$ & $51.42$ &  $59.06$ & $61.24$ & $61.56$ & $\textbf{63.10} \pm 0.14$\tabularnewline
    40 & $42.70$ & $46.40$ & $46.72$ & $47.14$ & $48.03$ & $54.54$ & $56.22$ & $59.94$ & $\textbf{60.02} \pm 0.15$\tabularnewline
    60 & $37.30$ & $40.58$ & $43.14$ & $43.80$ & $45.10$ & $52.36$ &  $52.84$ & $\textbf{54.92}$ & $54.41 \pm 0.49$\tabularnewline
    80 & $29.76$ & $33.58$ & $34.50$ & $33.46$ & $35.50$ & $40.00$ & $43.42$ & $45.62$ & $\textbf{46.51} \pm 0.20$\tabularnewline
       \bottomrule
    \end{tabular}}
     
\end{table*}

\subsection{Ablation study}
To understand better the benefit of each component to the final classification accuracy, we run an ablation study in Table~\ref{tab:ablacont}. We study the case of in-distribution noise only ($r_{in} = 0.4$), when out-of-distribution noise is present ($r_{in} = 0.2, r_{out} = 0.4$) and in the case of web noise (CNWL with $40\%$ web noise). We note that the selection of pseudo-labels using the pseudo-loss significantly improves the classification accuracy when training on with datasets presenting both ID, OOD or Web noise.
We aslso evaluate how important pseudo-label selection is when optimizing the confidence guided contrastive objective $l_{cont}$. We run PLS and apply label selection in the classification objective $l_{classif}$ but not in the contrastive objective $l_{cont}$. Noisy samples are corrected using the current consistency regularization guess but incorrect pseudo-labels are not filtered for $l_{cont}$ (they are for $l_{classif}$). Table~\ref{tab:ablacont} reports best accuracy results for CIFAR-100 corrupted with $40\%$ in-distribution (ID) or $20\%$ ID together with $40\%$ out-of-distribution (OOD) from ImageNet32 and the CNWL dataset with 40\% web noise. 
While we observe no major change for the ID corruption, a significant decrease in classification accuracy can be observed when keeping incorrect pseudo-labels in the contrastive objective when OOD or web noise is present (last row). For CIFAR-100 corrupted with 40\% OOD and 20\% ID noise, the accuracy benefits of training the contrastive objective are negated when compared to our noise correction baseline (row 2). We believe this motivates further research on the harmful impact OOD noise and incorrect pseudo-labels have when training on a web noisy dataset using a supervised contrastive objective.

\subsection{State-of-the-art label noise robust algorithms}
We propose to compare against the follwing state-of-the-art label noise robust algorithms: mixup (M)~\cite{2018_ICLR_Mixup} has shown to be a strong regularization naturally robust to label noise; MentorMix~\cite{2020_ICML_MentorMix} (MM) uses a student-teacher architecture to detect noisy samples before ignoring then; FaMUS~\cite{2021_CVPR_FaMUS} (FaMUS) is a meta-learning algorithm to detect label noise; Dynamic Bootstrapping~\cite{2019_ICML_BynamicBootstrapping} (DB) fits a beta mixture to the loss of training samples to detect noisy ones; S-model~\cite{2017_ICLR_Smodel} (SM) uses a noise adaptation layer optimized using an EM algorithm; DivideMix~\cite{2020_ICLR_DivideMix} (DM) uses an ensemble of networks to detect noisy samples; PropMix~\cite{2021_arXiv_propmix} (PM) only corrects the simplest of the noisy samples according to their training loss; ScanMix~\cite{2021_arXiv_Scanmix} (SM) corrects samples using a semantic clustering approach; Robust Representation Learning~\cite{2021_ICCV_RRL} cluster class prototypes and use a weighed average of neighbors labels to correct noisy samples; Multi Objective Interpolation Training~\cite{2021_CVPR_MOIT} (MOIT) train an interpolated contrastive objective and use neighbor label agreement to detect noisy samples. Regarding algorithms performing explicit in- and out-of-distribution noise detection,  EvidentialMix~\cite{2020_WACV_EDM} (EDM) fits a three component GMM to the evidential-loss~\cite{2018_NeurIPS_evidentialloss}; JoSRC~\cite{2021_CVPR_JoSRC} (JoSRC) uses the Jensen–Shannon divergence; Dynamic Softening for Out-of-distribution Samples~\cite{2022_WACV_DSOS} (DSOS) uses the collision entropy and Spectral Noise clustering from Contrastive Features~\cite{2022_ECCV_SNCF} (SNCF) clusters unsupervised features using OPTICS; Progressive Label Correction~\cite{2021_ICLR_featurenoise} (PLC) iteratively refine their noise detection under Bayesian guaranties, Peer-Learning~\cite{2021_ICCV_weblyfinegrained}, Co-teaching~\cite{2018_NIPS_CoTeaching},  PENCIL~\cite{2019_CVPR_PENCIL} co-train two networks and identify clean samples by voting agreement; SELFIE~\cite{2019_ICML_SELFIE} select low entropy noisy samples to be relabeled while discarding the rest.

\subsection{Synthetic corruption}
We first evaluate the capacity of PLS to mitigate in-distribution synthetic corruption. We run experiments on CIFAR-100 and compare against state-of-the-art algorithms in Table~\ref{tab:sotac100in}. To evaluate how much of the compared algorithms improvements come from an improve baseline accuracy over a superior noise correction, we also run the scenario where no noise is present. Because we do not use tricks such as unsupervised regularization and network ensembling, our algorithm presents a lower baseline when no noise is present but we achieve state-of-the-art results as soon as noise is introduced. This demonstrates the superiority of our approach when label noise is present in datasets.

\subsection{Out-of-distribution corruption}
Real world noisy data is often out-of-distribution~\cite{2022_WACV_DSOS}. We propose here to evaluate the performance of label noise robust algorithms on CIFAR-100 corrupted by a mixture between out-of-distribution images from ImageNet32 or Places365 and symmetric in-distribution noise. Table~\ref{tab:sotac100inout} reports our results and compare with state-of-the-art algorithms. We observe here that the pseudo-loss allows us to effectively deal with out-of-distribution images which are filtered out in the pseudo-loss since no corrected label can be guessed by the neural network.

\subsection{Controlled web noise}
We validate our approach using the controlled web corruptions proposed in the CNWL datasets trained at a resolution of $32\times 32$. We report results in Table~\ref{tab:sotamini32} and observe improvements over state-of-the-art algorithms including SNCF~\cite{2022_ECCV_SNCF} which uses self-supervised pre-training to detect label noise.

\begin{table*}[t]
    \caption{Hyperparameters used in the experiments~\label{tab:hyperparams}}
    \centering
    \global\long\def\arraystretch{0.9}%
    \resizebox{.7\textwidth}{!}{{{}}%
    \begin{tabular}{l>{\centering}c>{\centering}c>{\centering}c>{\centering}c>{\centering}c>{\centering}c>{\centering}c>{\centering}c>{\centering}c>{\centering}c>{\centering}c>{\centering}c>{\centering}c}
    \toprule
	Dataset & $r_{in}$ & $r_{out}$ & lr & epochs & Res & Net & GMM thresh & wd & warmup \tabularnewline
	\midrule
	\multirow{4}{*}{CIFAR-100} & 0.0 & 0.0 & 0.1 & 200 & 32 & PreRes18 & 0.95 & $5\times 10^{-5}$ & 30 \tabularnewline
	& 0.2 & 0.0 & 0.1 & 200 & 32 & PreRes18 & 0.95 & $5\times 10^{-5}$ & 30 \tabularnewline
	& 0.5 & 0.0 & 0.1 & 200 & 32 & PreRes18 & 0.95 & $5\times 10^{-5}$ & 30 \tabularnewline
	& 0.8 & 0.0 & 0.1 & 200 & 32 & PreRes18 & 0.5 & $5\times 10^{-5}$ & 30 \tabularnewline
	\midrule
	\multirow{4}{*}{CIFAR-100} & 0.2 & 0.2 & 0.1 & 200 & 32 & PreRes18 & 0.95 & $5\times 10^{-5}$ & 30 \tabularnewline
	& 0.2 & 0.4 & 0.1 & 200 & 32 & PreRes18 & 0.95 & $5\times 10^{-5}$ & 30 \tabularnewline
	& 0.2 & 0.6 & 0.1 & 200 & 32 & PreRes18 & 0.95 & $5\times 10^{-5}$ & 30 \tabularnewline 
	& 0.4 & 0.4 & 0.1 & 200 & 32 & PreRes18 & 0.95 & $5\times 10^{-5}$ & 30  \tabularnewline
	\midrule
	\multirow{4}{*}{CNWL} & 0.0 & 0.2 & 0.1 & 200 & 32 & PreRes18 & 0.95 & $5\times 10^{-5}$ & 1 \tabularnewline
	& 0.0 & 0.4 & 0.1 & 200 & 32 & PreRes18 & 0.95 & $5\times 10^{-5}$ & 1 \tabularnewline
	& 0.0 & 0.6 & 0.1 & 200 & 32 & PreRes18 & 0.95 & $5\times 10^{-5}$ & 1 \tabularnewline
	& 0.0 & 0.8 & 0.1 & 200 & 32 & PreRes18 & 0.95 & $5\times 10^{-5}$ & 1 \tabularnewline
	\midrule
	Web-aircraft & -- & -- & 0.003 & 110 & 448 & Res50 & 0.95 & $10^{-3}$ & 10 \tabularnewline
	Web-bird & -- & -- & 0.003 & 110 & 448 & Res50 & 0.95 & $10^{-3}$ & 10 \tabularnewline
	Web-car & -- & -- & 0.003 & 110 & 448 & Res50 & 0.95 & $10^{-3}$ & 10  \tabularnewline
    \bottomrule
    \end{tabular}}
\end{table*}

\subsection{Real world noise}
We conduct experiments on real world noisy datasets that have been directly crawled from the web with no human curation.  Table~\ref{tab:sotawebfg} reports results for the fine grained web datasets with results for other algorithms from Zeren~\etal~\cite{2022_CVPR_PNP}. Note that even if we use a single network to train and make a prediction, we get comparable or better results than ensemble or co-learning methods. We report results for algorithms that do not use the label softening strategy (LSR~\cite{2016_CVPR_rethinkingIncep}) as using this regularization technique is show in~\cite{2021_ICCV_weblyfinegrained} to provide a strong baseline performance boost independently of the noise correction capabilities of the algorithms that use it.

\subsection{Hyper-parameters}
Table~\ref{tab:hyperparams} reports the hyper-parameters used in all experiments.

\section{Conclusion}
This paper proposes a novel way to detect incorrect pseudo-label correction when dealing with label noise corruption. We use a state-of-the-art noise detection metric to detect noisy samples and guess their true label using a consistency regularization approach. The validity of the guessed true label is then evaluated using the pseudo-loss, which we show to be strongly correlated with pseudo-label correctness. Weight updates computed on pseudo-labels with a low probability of being correct are removed during training. We additionally propose to use an interpolated contrastive objective where correct pseudo-labels are used to learn inter class semantics while images with incorrect pseudo-labels are used in an unsupervised objective. We achieve state-of-the-art results on both synthetic and real world data.

\section*{Acknowledgments} 
This publication has emanated from research conducted with the financial support of Science Foundation Ireland (SFI) under grant number 16/RC/3835 - Vistamilk and 12/RC/2289\_P2 - Insight as well as the support of the Irish Centre for High End Computing (ICHEC).

{\small
\bibliographystyle{ieee_fullname}
\bibliography{egbib}
}

\end{document}


\title{Supplementary material for \\ PLS: Robustness to label noise with two stage detection}
\author{}

\maketitle

\thispagestyle{empty}

\section{Interpolated contrastive label}
\begin{figure*}[t]
\centering
\includegraphics[width=\linewidth]{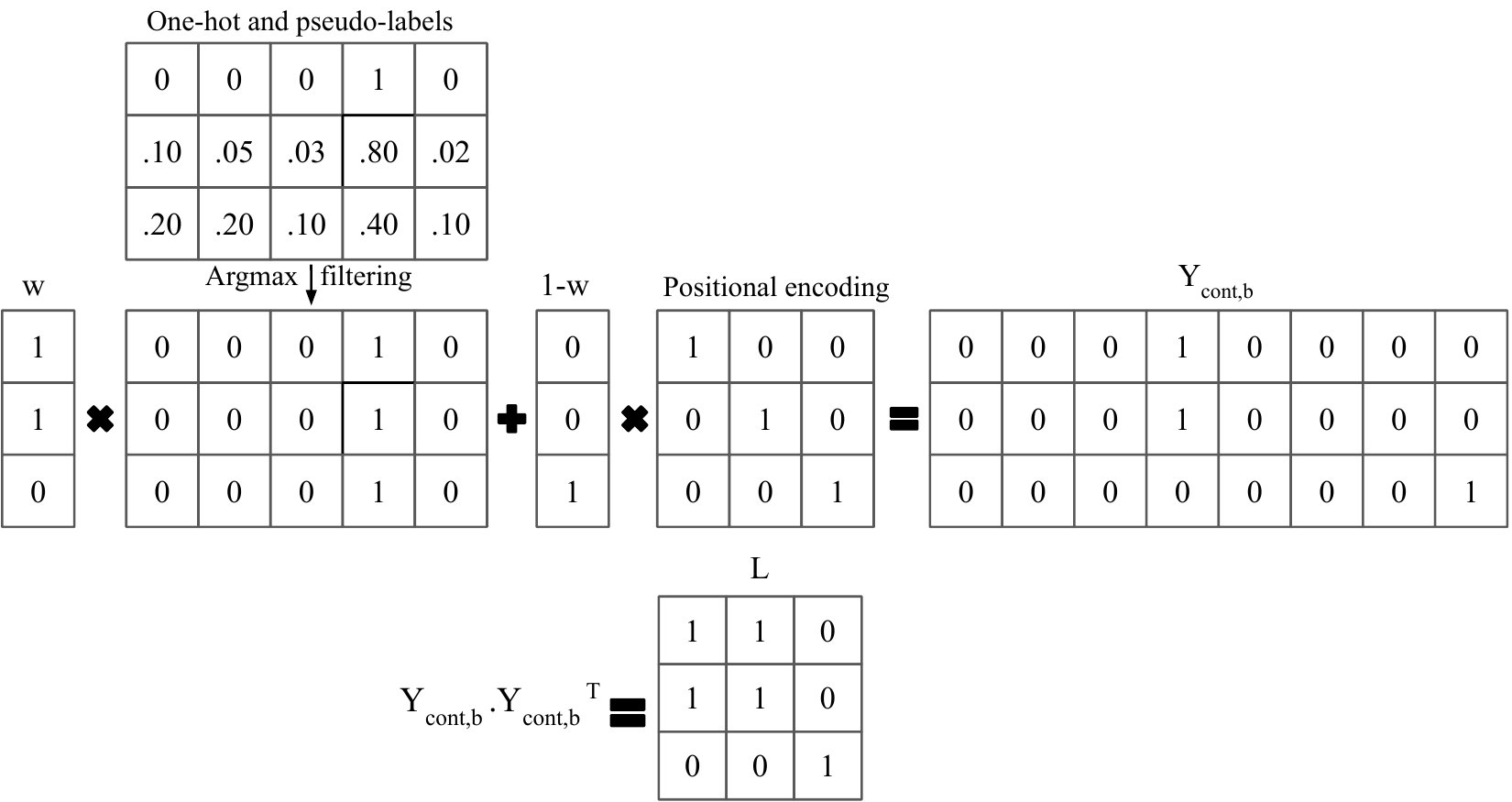}
\par
\caption{Toy example for the interpolated contrastive label between labeled and unlabeled with a mini-batch size b=$3$ and $5$ classes. Multiplications are computed row-wise and the + sign indicates a concatenation. In this example, the sample in the first row is detected clean, the second is noisy but reliably corrected and the third is noisy but unreliably corrected.~\label{fig:labelcont}}
\end{figure*}

Figure~\ref{fig:labelcont} illustrates the process for computing the contrastive label $L$ using the probability for a pseudo-label to be correct $w$, the guessed pseudo-labels and the positional encoding vector. Instead of simply ignoring the samples whose pseudo-label we can't reliably guess, we learn unsupervised features by enforcing the network to learn similar features between augmented views of a same image. For samples that are clean or whose pseudo-label we can rely on, we enforce the network to learn inter-image semantics between images of a same class.

\section{Hyper-parameter study}
Table~\ref{tab:ablamu} reports a study for values of $\mu$ (consistency regularization temperature) and $\gamma$ (contrastive loss temperature). We report the top-1 accuracy for different on CIFAR-100 corrupted with 40\% of synthetic in-distribution noise.

\section{Pseudo-loss selection vs confidence threshold}
A common strategy which has been used in the semi-supervised litterature~\cite{2020_arXiv_FixMatch,2021_NeurIPS_flexmatch} and to some extent in the label nosie litterature~\cite{2021_ICCV_RRL} is to only keep pseudo-label whose confidence is superior to a threshold. We compare this approach against our pseudo-loss selection in Table~\ref{tab:ablaconf} where we find that our pseudo-loss selection produces a better final validation accuracy. This is probably due to the fact that our selection computes a new threshold every epoch using an unsupervised Gaussian mixture. We believe that this allows to dynamically adapt to the learning state of the network as training progresses.

\section{PLS algorithm}
Algorithm~\ref{alg:algo} presents pseudocode for the PLS algorithm where we train a neural network $\phi$ robustly on a label noise dataset $\mathcal{D}$

\begin{table}[t]
    \caption{Ablation study on CIFAR-100 corrupted with $r_{in} = 0.4$.~\label{tab:ablamu}}
    \global\long\def\arraystretch{0.9}%
    \centering
    \resizebox{\linewidth}{!}{%
    \centering
    \begin{tabular}{l>{\centering}c>{\centering}c>{\centering}c>{\centering}c>{\centering}c>{\centering}c>{\centering}c>{\centering}c>{\centering}c>{\centering}c>{\centering}c>{\centering}c}
    \toprule
    $\mu$ & 0.1 & 0.15 & 0.2 & 0.25 & 0.3 & 0.5 \tabularnewline
    \midrule
    Top-1 val accuracy &  $77.86$ & $78.65$ & $78.45$ & $78.78$ & $78.45$ & $77.74$ \tabularnewline
    \midrule
    $\gamma$ & $0.5$ & $1$ & $1.5$ & $2$ & $2.5$ & $3$ \tabularnewline
    \midrule
    Top-1 val accuracy & $76.39$ & $78.54$ & $78.43$ & $78.33$ & $78.45$ & $77.99$ \tabularnewline
    
    \bottomrule
    \end{tabular}}
\end{table}
    
\begin{table}[t]
    \caption{Thresholds on pseudo-label confidence vs our selection. CIFAR-100 corrupted with $r_{in} = 0.4$.~\label{tab:ablaconf}}
    \global\long\def\arraystretch{0.9}%
    \centering
    \resizebox{\linewidth}{!}{%
    \centering
    \begin{tabular}{l>{\centering}c>{\centering}c>{\centering}c>{\centering}c>{\centering}c>{\centering}c>{\centering}c>{\centering}c>{\centering}c>{\centering}c>{\centering}c>{\centering}c}
    \toprule
    thresh & 0.8 & 0.9 & 0.95 & 0.98 & 0.99 & ours \tabularnewline
    \midrule
    Top-1 val accuracy & $77.40$ & $77.38$ & $77.18$ & $77.19$ & $77.05$ & $78.45$ \tabularnewline
    
    \bottomrule
    \end{tabular}}
\end{table}

\begin{algorithm*}[t]
\caption{PLS \label{alg:algo}}
\textbf{Input}: $\mathcal{D} = \left\{ \left(x_{i},y_{i}\right)\right\} _{i=1}^{N} $ a web noise dataset. $\phi$ a randomly initialized CNN and $g$ a linear projection to the contrastive space. \\
\textbf{Parameters}: $\alpha, \gamma, \mu, e_\text{warmup}, e_\text{max}, gmm_\text{thresh}$\\
\textbf{Output}: Trained neural network $\phi$
\begin{algorithmic}[1] 
\For{$e = 1, \dots e_\text{warmup}$} \Comment{Warmup}
\For{$t=1, \dots \mathit{batches}$}
\State Sample the next mini-batch $(x,y)$ from $\mathcal{D}$
\State $x_\text{mixed}, y_\text{mixed}$ = mixup(x, y, $\alpha$)
\State $l = \text{crossEntropy}(\phi(x_\text{mixed}), y_\text{mixed})$
\State $h = \text{updateNetworkWeights}(l)$
\EndFor
\EndFor
\State
\For{$e = e_\text{warmup}+1, \dots e_\text{max}$} \Comment{PLS correction}
\State isNoisy = detectNoise($\mathcal{D}$, $\phi$, $gmm_\text{thresh}$) \Comment{Detect the noise using the small loss and a GMM}
\For{$t=1, \dots \mathit{batches}$}
\State Two weakly augmented views $(x_1,y)$ and $(x_2,y)$ from $\mathcal{D}$
\State $pseudoLab =$ constReg($\phi(x_1)$, $\phi(x_2)$, $\gamma$) \Comment{Pseudo-label guess from Eq. 1}
\State $y[\text{isNoisy}] = pseudoLab[\text{isNoisy}]$ \Comment{Replace the labels of detected noise with the pseudo-labels}
\State Unaugmented view $(x,y)$ from $\mathcal{D}$
\State $pseudoLoss = \text{crossEntropy}(\phi(x), pseudoLab)$ \Comment{In practice, done in the detectNoise function}
\State $w = \text{GMM}(pseudoLoss)$ \Comment{Compute the probability to belong to the low loss mode of the pseudoLoss}
\State $x_\text{mixed}, y_\text{mixed}$ = mixup$(x_1, y, \alpha)$ \Comment{Mixup with corrected labels}
\State $l_\text{classif} = w \times \text{crossEntropy}(\phi(x_\text{mixed}), y_\text{mixed})$ \Comment{Weighed cross-entropy}
\State
\State $y = \text{oneHot}(y)$ \Comment{One-hot filtering}
\State $L = \text{contLabel}(y, w)$ \Comment{Compute the interpolated contrastive label Sec. 3.2.4}
\State Strong aug $X'$ from $\mathcal{D}$
\State $sims = (g(\phi(x_\text{mixed})) . g(\phi(X'))^T)/ \mu$ \Comment{Contrastive feats through projection}
\State $l_\text{cont} = $ contrastiveLoss(sims, L) \Comment{From Eq. 5}
\State
\State $h = \text{updateNetworkWeights}(l_\text{classif}+l_\text{cont})$
\EndFor
\EndFor
\State \textbf{return} $\phi$ \Comment{Robustly trained network}
\end{algorithmic}
\end{algorithm*}

{\small
\bibliographystyle{ieee_fullname}
\bibliography{egbib}
}